\documentclass[10pt,twocolumn,letterpaper]{article}

\usepackage[pagenumbers]{iccv} 

\usepackage{graphicx}
\usepackage{booktabs}
\usepackage{multirow}

\definecolor{iccvblue}{rgb}{0.21,0.49,0.74}
\usepackage[pagebackref,breaklinks,colorlinks,allcolors=iccvblue]{hyperref}

\def\paperID{*****} 
\def\confName{ICCV}
\def\confYear{2025}

\title{Taming Modern Point Tracking for Speckle Tracking \\Echocardiography via Impartial Motion}

\author{
Md Abulkalam Azad$^{1,2}$ \quad
John Nyberg$^{1}$ \quad
Håvard Dalen$^{1,2}$ \quad
Bjørnar Grenne$^{1,2}$ \quad\\
Lasse Lovstakken$^{1}$ \quad
Andreas {\O}stvik$^{1,2,3}$ \\
$^1$Norwegian University of Science and Technology, Trondheim, Norway \\
$^2$Clinic of Cardiology, St. Olavs Hospital, Trondheim, Norway \\
$^3$SINTEF Digital, Trondheim, Norway \\
{\tt\small \{md.a.azad, andreas.ostvik\}@ntnu.no}
}

\begin{document}
\maketitle
\begin{abstract}
Accurate motion estimation for tracking deformable tissues in echocardiography is essential for precise cardiac function measurements. While traditional methods like block matching or optical flow struggle with intricate cardiac motion, modern point tracking approaches remain largely underexplored in this domain. This work investigates the potential of state-of-the-art (SOTA) point tracking methods for ultrasound, with a focus on echocardiography. Although these novel approaches demonstrate strong performance in general videos, their effectiveness and generalizability in echocardiography remain limited. By analyzing cardiac motion throughout the heart cycle in real B-mode ultrasound videos, we identify that a directional motion bias across different views is affecting the existing training strategies. To mitigate this, we refine the training procedure and incorporate a set of tailored augmentations to reduce the bias and enhance tracking robustness and generalization through impartial cardiac motion. We also propose a lightweight network leveraging multi-scale cost volumes from spatial context alone to challenge the advanced spatiotemporal point tracking models.
\begin{figure}[htpb]
\centering
\includegraphics[width=\linewidth]{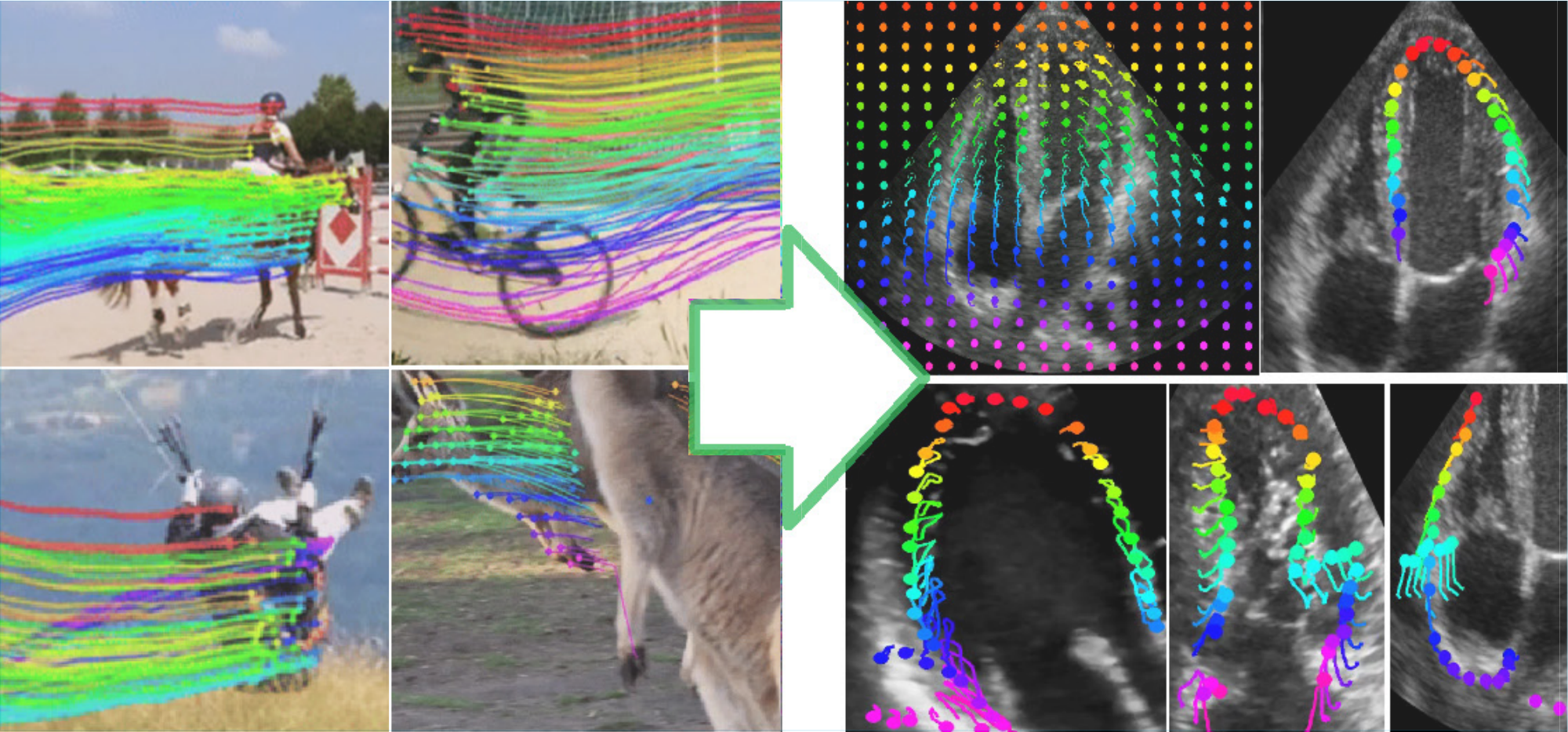}
\caption{From point tracking of large, unidirectional motion to speckle tracking of smaller, cyclic cardiac motion.} 
\label{fig:paradigm}
\end{figure}
Experiments demonstrate that fine-tuning with our strategies significantly improves models' performances over their baselines, even for out-of-distribution (OOD) cases. For instance, EchoTracker boosts overall position accuracy by 60.7\% and reduces median trajectory error by 61.5\% across heart cycle phases. Interestingly, several point tracking models fail to outperform our proposed simple model in terms of tracking accuracy and generalization, reflecting their limitations when applied to echocardiography. Nevertheless, clinical evaluation reveals that these methods improve global longitudinal strain (GLS) measurements, aligning more closely with expert-validated, semi-automated tools and thus demonstrating better reproducibility in real-world applications. All fine-tuned models and code will be released at: https://github.com/***/*** upon acceptance.
\end{abstract}

\begin{figure}[htpb]
\centering
\includegraphics[width=0.98\linewidth]{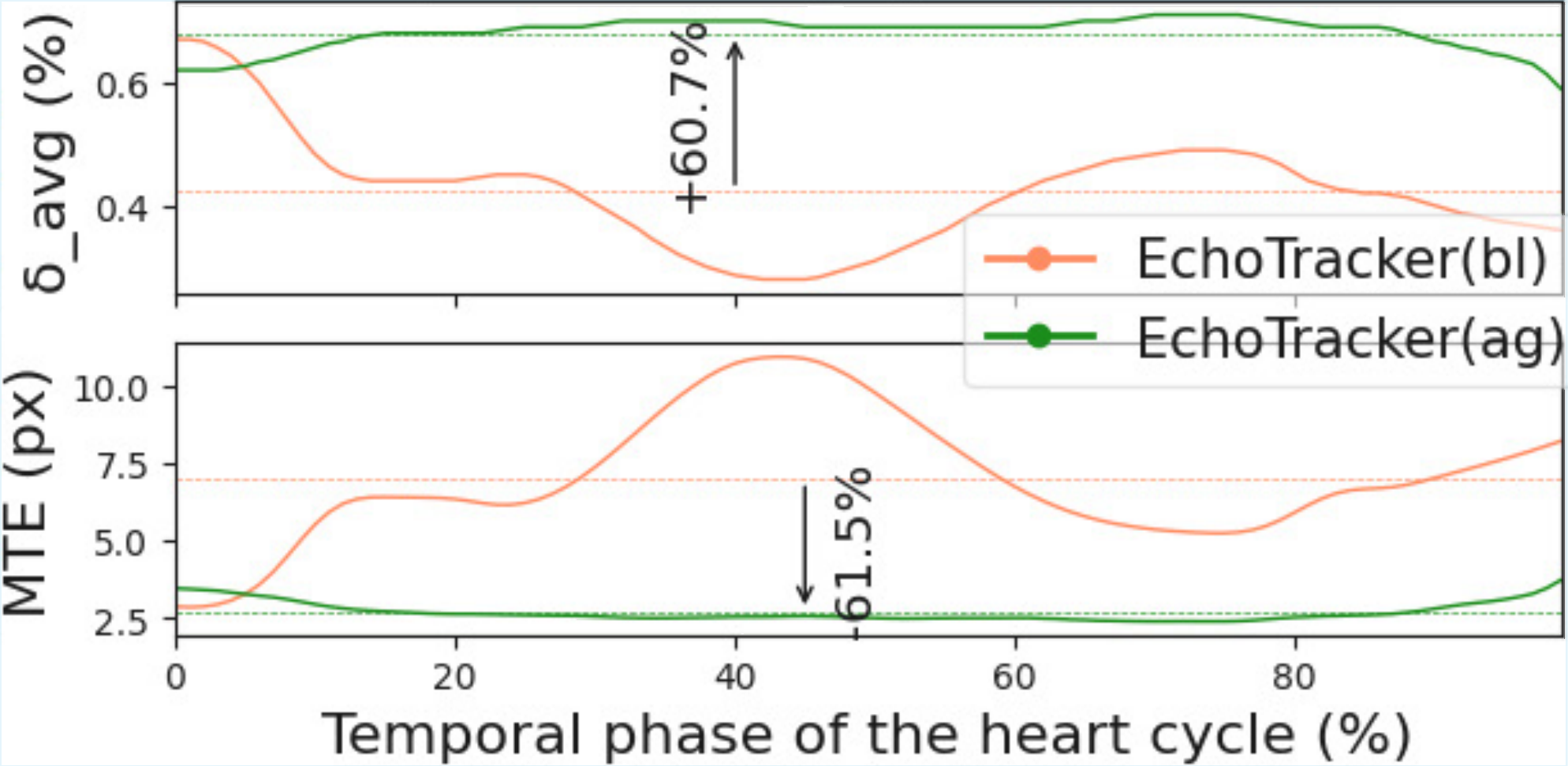}
\caption{The re-fine-tuned EchoTracker(ag) demonstrates greater temporal invariance by improved positional accuracy (top) and reduced trajectory error (bottom) across all temporal phase initializations within the heart cycle compared to the baseline (bl).} 
\label{fig:echotracker_perf}
\end{figure}
\section{Introduction}
\label{sec:intro}
Speckle tracking (ST) enables automatic tracking of regions of interest through ultrasound (US) images. An optimal ST approach for detecting speckle motion and deformation in US must adhere to fundamental principles. Specifically, the speckle pattern, defined by the spatial distribution of greyscale values, is expected to uniquely identify each myocardial segment and follow the motion~\cite{d2007principles}. However, practical factors like reverberation, translation, rotation, deformation, out-of-plane motion, frame rate, and different artifacts can cause speckle decorrelation, challenging this assumption. Methods like block or feature matching rely on coherent high-quality US data and only capture similarities without considering cardiac motion dynamics~\cite{voigt2015definitions}. Numerous alternative approaches, including optical flow, have been proposed to address these issues. Nevertheless, they face difficulties with long-range tracking due to the absence of temporal priors beyond consecutive frames and fail to account for deformability and appearance changes, causing drift~\cite{angelini2006review,joos2018high,poree2018dual}. These challenges persist even in learning-based approaches~\cite{ostvik2021myocardial,ostvik2018automatic,salte2021aistrain}. 

Recently, a new research direction in deep learning, known as \emph{point tracking}, has emerged as a new paradigm for long-range motion estimation in videos~\cite{doersch2022tap,doersch2024bootstap,doersch2023tapir,harley2022particle,karaev2024cotracker3,karaev2023cotracker,zheng2023pointodyssey}. This approach enables tracking of arbitrary points across both rigid and deformable objects over time, as shown in Fig.~\ref{fig:paradigm} (left). Unlike traditional methods, it offers better handling of long-range dependencies and is naturally adaptable to challenging scenarios like speckle decorrelation in US sequences (Fig.~\ref{fig:paradigm}, right). While various methods, including CoTracker3~\cite{karaev2024cotracker3} and TAPIR~\cite{doersch2023tapir},  have been proposed for general computer vision tasks, their applicability to medical US remains largely unexplored. To address this gap, Azad et al.~\cite{azad2024echotracker} introduced EchoTracker, a lightweight coarse-to-fine architecture specifically designed for tracking myocardial points in US videos. Although the architectural design was tailored to US characteristics, the augmentations and training strategies were adopted directly from Zheng et al.~\cite{zheng2023pointodyssey}, without adaptation to cardiac motion. Additionally, unlike SOTA point tracking methods, they restricted query points to be selected only from the first frame of the heart cycle, which may hinder optimal tracking and limit clinical usability. As shown in Fig.~\ref{fig:echotracker_perf}, baseline EchoTracker (orange) shows a gradual decrease in average position accuracy and an increase in median trajectory error as query points shift from the beginning (frame 0) to later temporal phases of the heart cycle. While similar phenomena are observed in other methods, the effect is more intense in EchoTracker. This indicates that, like other SOTA methods, EchoTracker struggles with the motion bias and lacks robustness across different cardiac phases and views. Moreover, although models fine-tuned following Azad et al.~\cite{azad2024echotracker} can capture overall tissue movement, their precision remains suboptimal, with limited temporal invariance causing performance fluctuations when initialized at different time frames along the same trajectories.

In this work, we explore novel point tracking methods for speckle tracking echocardiography (STE), integrate them into a deep learning pipeline~\cite{ostvik2021myocardial,ostvik2018automatic} to automatically compute GLS, and compare results against a commercially available system. Additionally, since these methods leverage long-range temporal priors, we aim to overcome one of the most overlooked issues in STE—initializing models at a fixed time frame within the heart cycle for optimal tracking. However, we argue that such fixed initialization may struggle to capture anomalous heart motions caused by various cardiac conditions and variations in heart size or shape. To summarize, we first adopt several SOTA point tracking methods from general computer vision, fine-tune them following the training protocol of Azad et al.~\cite{azad2024echotracker} to obtain their baselines for this study, and evaluate their efficacy in echocardiography. To benchmark their performance, we propose a simple multi-scale speckle tracking network, \emph{SpeckNet}, which performs feature matching for query points across all frames purely in spatial resolution at different scales. The goal is to assess how well these advanced point tracking models perform relative to a straightforward learning-based feature matching approach. Secondly, we analyze cardiac motion in our datasets, identify a directional motion bias, and address it by incorporating tailored motion augmentations and refining the training strategy to improve robustness, generalizability, and temporal consistency. Thirdly, to validate the effectiveness of our combined strategies, we fine-tune the models again and evaluate their tissue tracking performance and temporal initialization consistency against their baselines across diverse cardiac US datasets, including OODs. Finally, to assess clinical relevance, we compute GLS using both baseline and re-fine-tuned models and compare the results with expert reference measurements by a semi-automatic commercial system from GE HealthCare. All fine-tuned models are to be publicly released to support the research community and foster further advancements.



\section{Related Work}
STE has evolved from classic feature or block-matching approaches to optical flow methods~\cite{trahey1988angle,suhling2005myocardial,meunier1988local,mailloux1987computer,horn1981determining}. Traditional block-matching techniques involve dividing the image into blocks and identifying the best match in consecutive frames, which can be computationally intensive and less precise in detecting subtle tissue motion due to speckle decorrelation over time within the sequence. Traditional optical flow methods, on the other hand, estimate pixel-wise motion vectors based on intensity changes between consecutive frames. These approaches rely on assumptions like brightness constancy and spatial coherence, simplifying motion estimation through local information and gradient-based techniques. Although effective in many scenarios, they often struggle with complex motion patterns, occlusions, and large displacements, leading to inaccuracies, especially with rapid or non-rigid movements. Moreover, the absence of temporal context beyond adjacent frames introduces cumulative errors over time. Also, dense pixel-wise computations are computationally intensive, particularly when tracking is required only for specific regions rather than the entire image. Although many learning-based optical flow methods~\cite{dosovitskiy2015flownet,ilg2017flownet,sun2018pwc} have been developed to overcome the limitations of traditional approaches by better adapting to complex motion patterns and showing improved robustness and generalization across various scenarios, they still face similar challenges due to the underlying assumption that neighboring pixels share identical motion. Nonetheless, some studies have applied these learning-based optical flow methods~\cite{deng2022myocardial,evain2022motion,ostvik2021myocardial} to facilitate myocardial motion tracking.

\textbf{Point Tracking}, a new research in modern deep learning techniques, has emerged over the past couple of years to overcome the limitations of learning-based optical flow~\cite{harley2022particle,poree2018dual,doersch2022tap,doersch2023tapir,karaev2023cotracker}. Unlike optical flow, it focuses on estimating the motion of specific query points on an object or instance, rather than computing motion vectors for the entire spatial field. These methods typically compute a cost volume by correlating the feature vector of each query point with its local neighbours within a region of interest. To incorporate temporal information, strategies like 1D convolutional neural networks with temporal kernels or transformers with spatial and temporal attention are employed. These techniques iteratively update the query features, refine the trajectory over time and improve motion estimation with each iteration. Particular works include Harley et al.~\cite{harley2022particle}'s adaptation of Sand and Teller’s Particle Video~\cite{sand2008particle} by combining sparse feature matching with dense optical flow for pixel-level tracking. While their particle trajectories improved robustness to occlusion and yielded smoother motion, they struggled with long-range tracking — a limitation later addressed in PIPs++~\cite{zheng2023pointodyssey} by proposing multi-template updates to query features. Concurrently, Doersch et al. introduced TAP-Net for motion estimation using spatially correlated cost volumes~\cite{doersch2022tap}, extended it with TAPIR’s two-stage matching and refinement~\cite{doersch2023tapir}, and later improved it further with self-supervised BootsTAPIR~\cite{doersch2024bootstap}. Furthermore, transformer-based approaches advancing point tracking include LocoTrack~\cite{cho2024local}, CoTracker~\cite{karaev2023cotracker}, DINO-Tracker~\cite{tumanyan2024dino}, TAPTR~\cite{li2024taptr}, and the more recent Track-On~\cite{Aydemir2025ICLR}. CoTracker pioneered the use of transformer architecture for point tracking, with subsequent methods introducing various enhancements. Later, Karaev et al. unified these advancements into a simplified architecture, CoTracker3~\cite{karaev2024cotracker3}, offering both online and offline tracking. However, their adoption in medical imaging, particularly US, remains unexplored. EchoTracker~\cite{azad2024echotracker}, MyoTracker~\cite{chernyshov2025low}, and PIPsUS~\cite{chen2024pipsus} focus on 2D myocardial and neck/oral US tracking, respectively, while NeuralCMF~\cite{shen2024continuous} targets 3D motion tracking in echocardiography. Therefore, while numerous options exist, we choose only PIPs++, TAPIR, BootsTAPIR, and CoTracker3 — representing the leading models with diverse architectures — alongside EchoTracker for our study to reduce computational hours and keep the carbon footprint limited.

\section{Datasets}
\begin{figure}[h]
\centering
\includegraphics[width=0.95\linewidth]{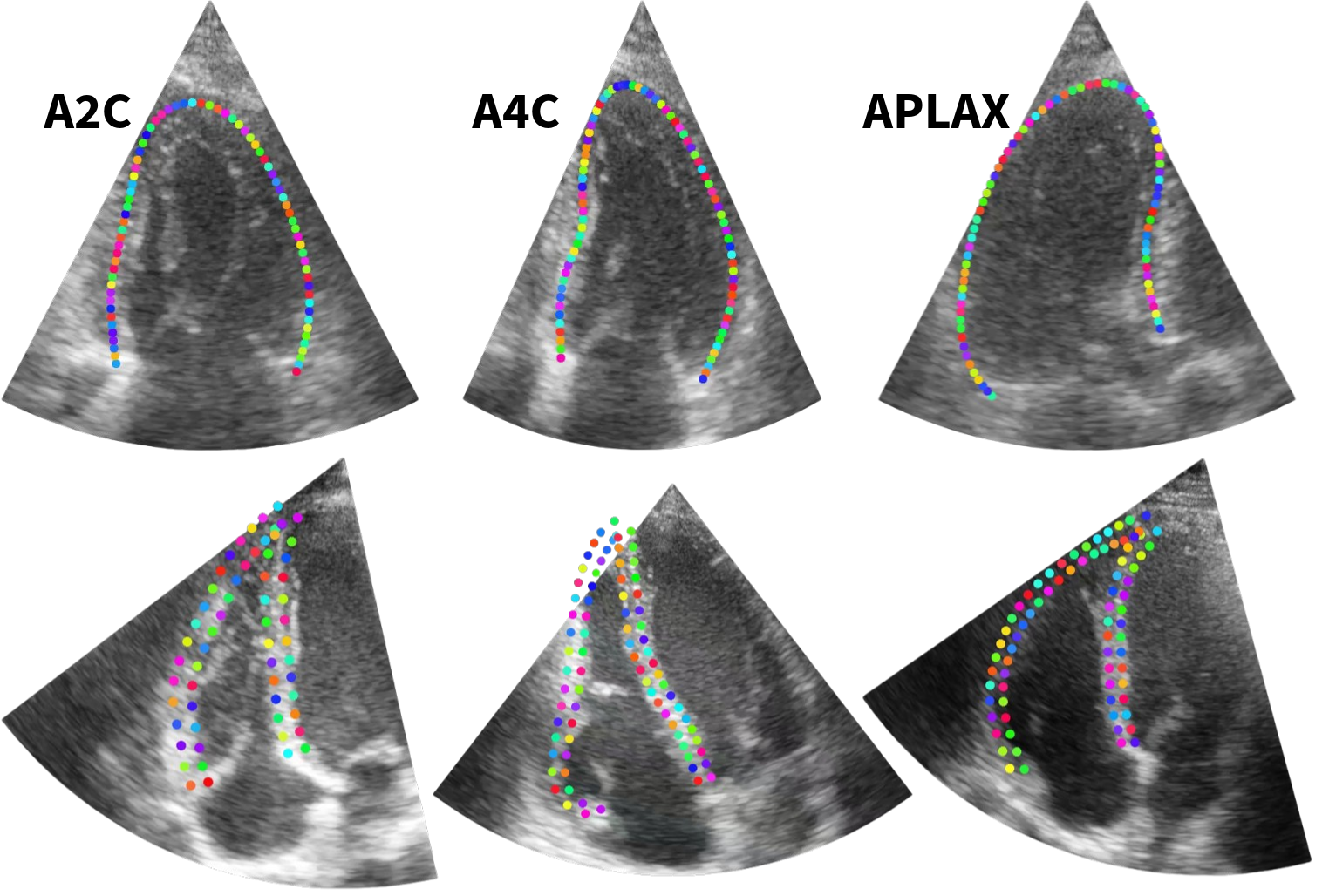}
\caption{Top row displays LV-focused data samples from three distinct views: apical two-chamber (A2C), apical four-chamber (A4C), and apical long-axis (APLAX). The second row presents RV-focused samples acquired from a single view (A4C).} 
\label{fig:ds}
\end{figure}

Since one of the contributions of this work lies in reference motion analysis to design an effective training procedure for point tracking methods in ST, we first introduce the available datasets and their characteristics. Each training sample consists of a video with \( T \) frames, where each frame contains \( N \) points defined by \((x, y)\) coordinates, forming \( N \) trajectories. Due to the scarcity of real-world datasets with ground-truth trajectories, most methods rely on sim-to-real transfer learning, where models are trained on simulated US data to estimate motion in real data~\cite{evain2022motion}. This work uses real-world patient datasets collected with informed consent, using GE Vivid E95 ultrasound systems with phased-array probes. Trajectories were initially generated by selecting regions of interest, such as the left or right ventricular (LV or RV) myocardium, and applying a traditional tracking algorithm in a semi-supervised manner. These trajectories were then refined by clinical experts ($>$ 20 years experience), who performed rigorous quality assessments, excluding points that did not reliably track myocardial tissue. Fig.~\ref{fig:ds} shows the first frames of selected video sequences with annotated myocardial points and the characteristics of datasets are summarized in Table~\ref{tab_ds}. Although resolutions vary across datasets, all samples are uniformly resized to \( 256 \times 256 \) for consistent experiments. Each dataset captures LV myocardium wall motion with centerline point trajectories covering a complete cardiac cycle, from end-diastole (ED) to end-systole (ES) and back to ED. The only exception is dataset E, which focuses on the RV myocardium wall with two edge-line trajectories and serves as an OOD dataset for our evaluation.

\begin{table}[ht]
\centering
\caption{Overview of ultrasound point tracking datasets (DS.), including the number of patients (Pts.), videos (Vids.), average number of point trajectories (Trajs.) and frames per video with minimum and maximum in parentheses, and pathology types.}
\label{tab_ds}
\setlength{\tabcolsep}{1.0pt} 
\begin{tabular}{lccccccccc}
\hline
 {\textbf{DS}}& \textbf{\#Pts.} & {\textbf{\#Vids.}}& \textbf{\#Trajs.} & \textbf{\#Frames} & \textbf{Pathology}\\
\hline
 D & 371 & 1112 & 75 (47-137) & 82 (44-185) & Mix(MI,  HF, AF)\\
 E & 701 & 735 & 61 (46-82) & 82 (37-161) & Healthy \\
 T & 643 & 1922 & 66 (45-111) & 84 (46-151) & Healthy\\
\hline
\end{tabular}
\\[0.1cm]
\raggedright
{\scriptsize HF: Heart failure, AF: Atrial fibrillation, MI: Myocardial infarction}
\end{table}

\section{Methods}
\label{sec:method}
\subsection{Myocardium Tissue Tracking}
Given the variety of tracking concepts in computer vision (e.g., object, pose, point, and face tracking), as well as different configurations within point tracking, we provide an explicit definition of myocardium tissue tracking through points for our problem. Let the model \(\psi\) take as input a cardiac ultrasound video \( V = \{ f_t \in \mathbb{R}^{H \times W} \} \), where \(f_t\) represents each frame with height \(H\) and width \(W\) for \( t = 0, 1, 2, \ldots, T-1 \). The input also includes a set of pixel coordinates \(0 \leq (x, y) < H/W\) representing query points initialized on any frame \(q\)  within the heart cycle, denoted as \( p_q = \{ (x_q^n, y_q^n) \} \) for \( n = 0, 1, 2, \ldots, N-1 \). The query frame \(q\) must remain constant for all query points, ensuring that they originate from a single frame. Afterwards, the model should analyze the underlying motion and output the pixel locations of the query points in all other frames. To reduce ambiguity between the adopted methods, we modify or utilize them to adhere to this uniform definition. However, since \( p_q \) is provided as input, it is excluded from the output when evaluated. The following equation summarizes the problem where \(q\) is a constant number, representing the query frame:
\begin{equation}
    \psi(V, p_q) = \{p_t \in (x_t^n, y_t^n)\}
\end{equation}

\subsection{SpeckNet}
To challenge the point tracking methods that leverage advanced strategies—such as multi-scale spatiotemporal features and iterative refinement for estimating fine trajectories—we introduce a simple speckle tracking spatial network, named \emph{SpeckNet}. As illustrated in Fig.~\ref{fig:specknet}, SpeckNet takes a high-resolution video $V$ with $H = W = 512$ as input to retain as much spatial detail from the original data as possible. To manage computational demands within limited memory, we reduce the spatial dimensions using a non-learnable pixel unshuffling operation~\cite{shi2016real} with a downsampling scale factor of 2. This transforms the input shape from $\mathbb{R}^{T \times H \times W}$ to $\mathbb{R}^{T \times H/2 \times W/2 \times 4}$, effectively preserving spatial information into channels without the losses typically occurred with standard downsampling. Next, the input is passed through a pruned ResNet with four block groups, similar to the architecture used in TAP-Net~\cite{doersch2022tap}, where each block group contains two residual sub-blocks. For simplicity, only the last three block groups are illustrated in Fig.~\ref{fig:specknet}. At each query point location, bilinear interpolation is used to extract feature vectors from the output feature maps of these three block groups. Cosine similarity is then computed between each feature vector and the corresponding feature maps across all frames, resulting in a cost volume per ResNet block. Specifically, we obtain three cost volumes of different spatial resolutions: $C_3 \in [-1, 1]^{T \times H/16 \times W/16}$, $C_2 \in [-1, 1]^{T \times H/8 \times W/8}$, and $C_1 \in [-1, 1]^{T \times H/8 \times W/8}$. To enhance matching fidelity, we first scale the intensity range from $[-1, 1]$ to $[0, 1]$ and apply a power operation $C^{'} = C^{\gamma}$ with $\gamma = 15$, which amplifies high similarity scores toward 1 and suppresses lower scores toward 0 — thereby increasing contrast in the cost volumes. These cost volumes are then upsampled by interpolation to the highest resolution and combined via element-wise multiplication. This strategy leverages complementary strengths:
\begin{figure}[htpb]
\centering
\includegraphics[width=\linewidth]{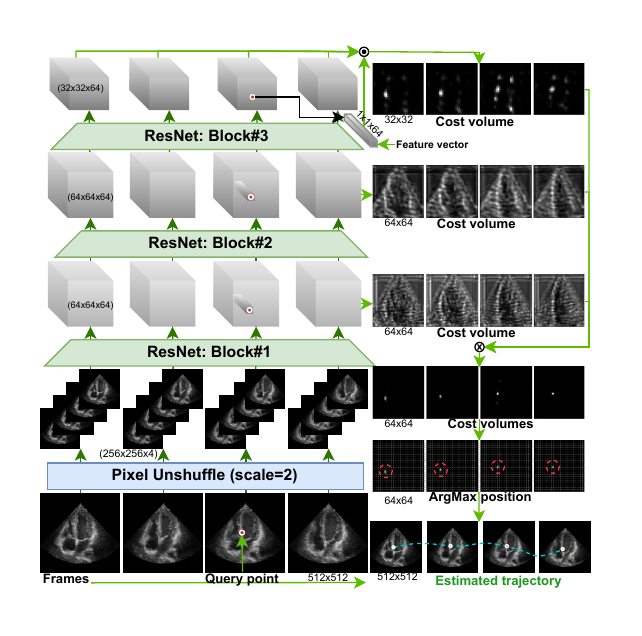}
\caption{Overview of the SpeckNet architecture. Feature vector extraction and cost volume computation steps are shown only for one block and omitted from others to reduce visual complexity.} 
\label{fig:specknet}
\end{figure}
\begin{figure*}[h]
\centering
\includegraphics[width=\linewidth]{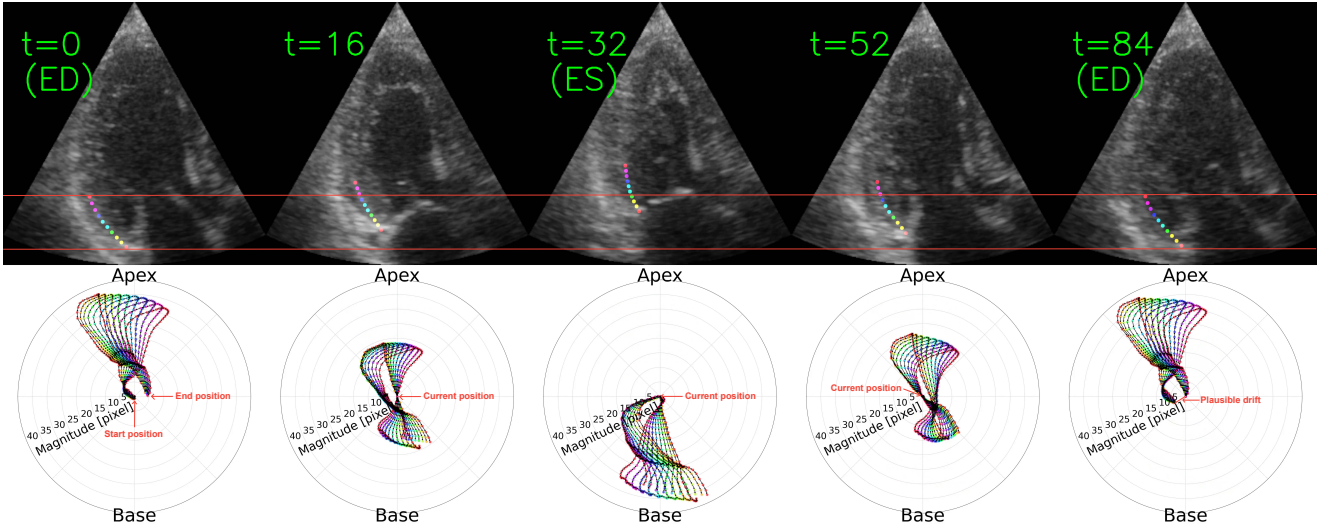}
\caption{Motion distribution in polar representation with displacement vectors (bottom) illustrates the magnitude and direction of tracking points (top) relative to their positions in the current frame at selected temporal phases within the cardiac cycle.} 
\label{fig:magnitude}
\end{figure*}
coarse-scale volumes offer robust but imprecise matches, while fine-scale volumes provide detailed but noisy signals. Their intersection via multiplication cancels out spurious matches and reinforces consistent ones, yielding a more refined cost volume. The use of two same sized high-resolution volumes was guided by an analysis suggesting improved tracking performance, potentially due to an effect analogous to multi-head self-attention~\cite{vaswani2017attention}. Finally, a soft argmax operation is applied within a five-pixel radius of the hard argmax location to estimate a smooth trajectory.

\subsection{Cardiac Motion Dynamics}
\begin{figure*}[htbp]
\centering
\includegraphics[width=\linewidth]{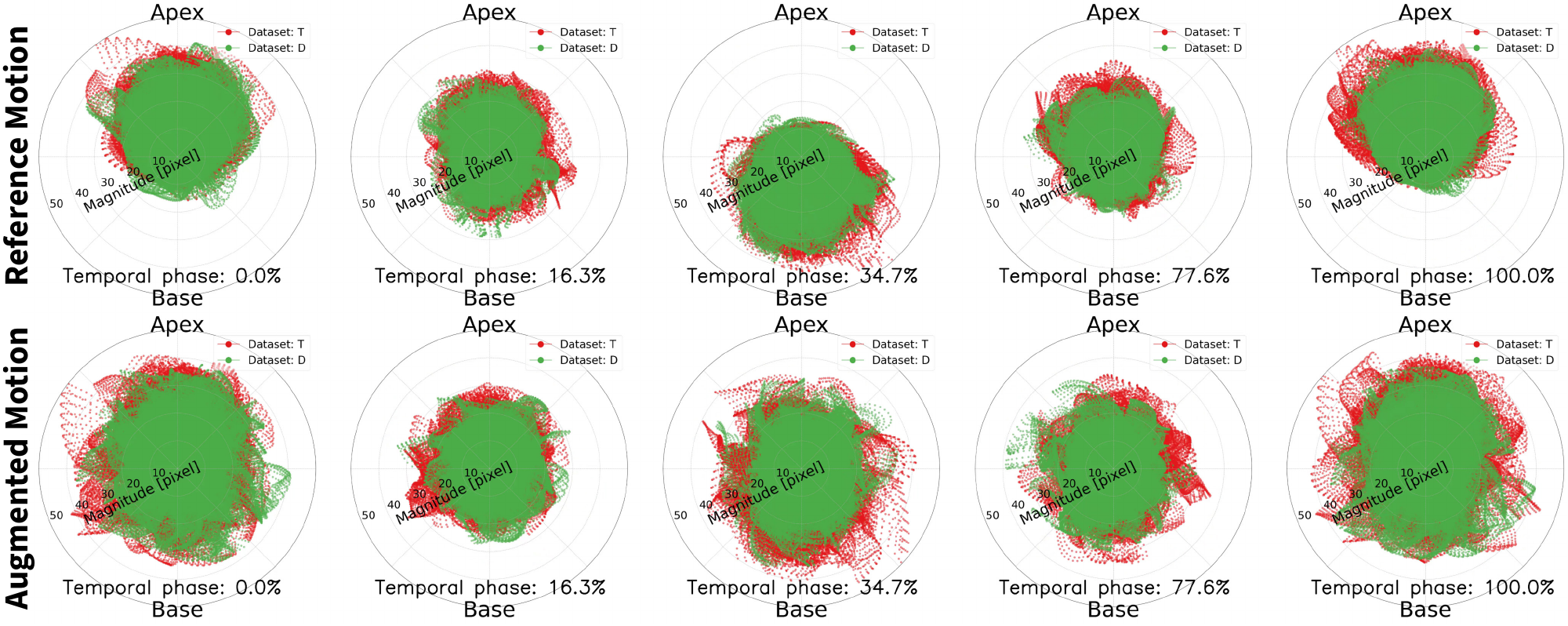}
\caption{Motion distribution in polar representation across temporal phases (left to right) of the cardiac cycle. The top row illustrates the original motion distribution for representative subsets from datasets D and T, while the bottom row shows the corresponding distributions after applying a set of affine transformations to the videos and their associated motion trajectories.} 
\label{fig:motion_dist}
\end{figure*}

Cardiac motion is dynamic and complex, driven by contraction, relaxation, and hemodynamic forces~\cite{pollok2023physiology}. Unlike general videos with large and unidirectional motion, cardiac motion is small and cyclic (Fig.~\ref{fig:paradigm}). To precisely analyze small motions, we convert the reference Cartesian points $(x_t^n, y_t^n)$ for each of the $N$ trajectories over $T$ frames into polar coordinates centered at their current frame locations $(x_c^n, y_c^n)$. The radius and angle are calculated by the following equations:
\begin{equation}
r^n_t = \sqrt{(x^n_t - x^n_c)^2 + (y^n_t - y^n_c)^2}
\end{equation}
\begin{equation}
    \theta^n_t = \arctan2(y^n_t - y^n_c, -(x^n_t - x^n_c))
\end{equation}
This anatomical projection captures motion or displacement of the entire trajectory within the heart cycle relative to the current frame, as illustrated in Fig.~\ref{fig:magnitude}, which visualizes 10 trajectories for points along the basal posterior wall for five selected frames. At \(t=0\) (ED), the polar trajectories indicate that the highest relative magnitudes (approximately 38 pixels) for the corresponding points are toward the apex from their starting positions and will end up at a slightly different position after completing the cycle due to drift. At \(t=32\) (ES), those points reach their peaks, and almost all motions — both forward and backward — are directed toward the base from the current position. At the final frame $t=84$, those points return to their end positions. This observation indicates that cardiac motion in the selected sample is strongly biased toward vertical movement, with minimal horizontal displacement across temporal phases. Similarly, we analyze motion patterns across all samples in our datasets and visualize a representative subset of D and T in Fig.~\ref{fig:motion_dist} (top row) for an accumulated illustration. An animation covering the full datasets is included in the supplementary material. The illustration suggests that the motion pattern in the entire dataset is consistent with the bias of vertical movement as shown in Fig.~\ref{fig:magnitude}, and also the distribution is uneven across temporal phases. Hence, training on such motion patterns may limit a point tracking model’s generalizability to other cardiac views, diverse populations, OOD samples, or scenarios involving probe movements during acquisition by less experienced clinicians. To mitigate these issues, we propose applying a set of affine transformations (detailed in the next section) to both the videos and corresponding trajectories.

\subsection{Augmentation and Training}
To enhance the robustness and generalizability of point tracking methods, it is essential for models to learn from a wide range of cardiac motion patterns, ideally derived from a diverse population. However, aggregating such a comprehensive dataset is often impractical. Therefore, we aim to eliminate the motion bias discussed in the previous sub-section by augmenting the training datasets with a set of affine transformations, making them more representative of diverse motion patterns across vendors, populations, age groups, and other variations. The adopted transformations include scaling, translation, shearing, rotation, and horizontal/vertical flipping, each applied with a probability of $p = 0.5$. While the parameters of these transformations vary across samples, they remain consistent within a given video to preserve temporal motion coherence. These augmentations simulate diverse cardiac motion patterns by affecting both the magnitude and direction of motion trajectories. Specifically, \emph{scaling} can mimic larger or smaller displacements, representing variability in heart size and function. \emph{Translation} addresses variability in myocardial location across different views, enhancing robustness to positional shifts. \emph{Shearing} introduces directional distortion, converting predominantly vertical motion into diagonally skewed trajectories. \emph{Horizontal} and \emph{vertical flipping} simulate mirrored cardiac views, allowing, for example, models trained on LV data to generalize to RV. Finally, random \emph{rotations} up to ±120 degrees can reorient vertical motion into horizontal or oblique directions. This enables the model to handle real-world conditions such as probe rotation or non-standard views, common in point-of-care ultrasound acquired by non-experts.
\begin{figure*}[h]
\centering
\includegraphics[width=\linewidth]{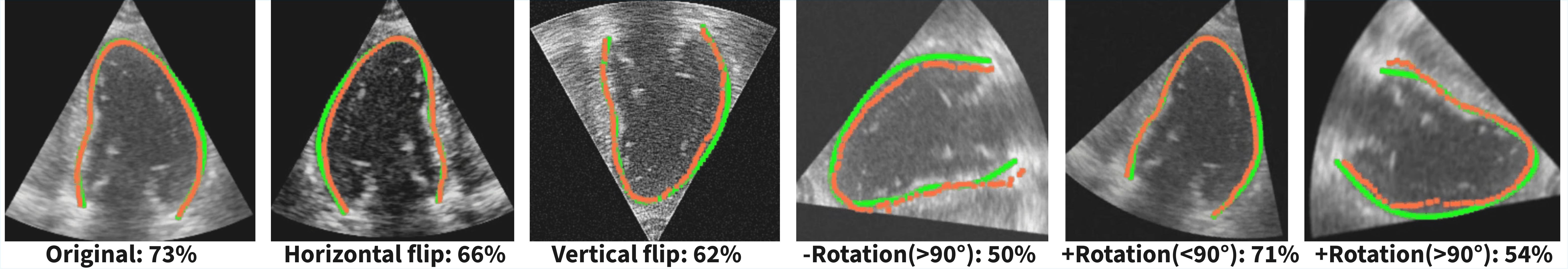}
\caption{Examples showing the effect of affine transformations as perturbations on the tracking performance of EchoTracker trained without the proposed augmentations. The model achieves an average positional accuracy of 73\% on the original (unperturbed) sample, which deteriorates under various transformations. Green tracks indicate the reference trajectories, while orange represents the estimation.} 
\label{fig:affine_aug}
\end{figure*}
Additionally, we apply image augmentations such as motion blurring, sharpening, embossing, brightness contrast adjustments, hue saturation, image compression, and Gaussian noise with probabilities \(p<0.5\) to enhance adaptability to speckle noise and artifacts. Fig.~\ref{fig:affine_aug} shows how combined augmentations affect tracking of the baseline EchoTracker. 

For optimal temporal initialization, the common practice is to use the frame (temporal phase) with the smallest overall displacement throughout the cardiac cycle. By analyzing the motion distribution animation of full datasets — examples of which are shown in Fig.~\ref{fig:motion_dist} (top row) — we identify two potential candidates: approximately 25\% and 75\% temporal phases of the cardiac cycle. Since cardiac motion is more rapid around the 25\% phase (ejection) compared to the 75\%, we hypothesize that initializing at 75\%, corresponding to diastasis, offers a more stable reference point. We validate this hypothesis through experiments presented in the next section. However, we employ this optimal initialization only during inference, as training exclusively from this phase would bias the models toward tracking small motions and reduce their exposure to larger, more complex displacements. Similarly, initializing solely from phases with large motion would introduce the opposite bias. Therefore, to promote temporal invariance, we randomly initialize the models at different temporal phases for each sample during training. Fig.~\ref{fig:motion_dist} also indicates that overall motion in healthy patients (T) remains consistently larger across all phases of the heart cycle than in sick patients (D). Beyond the clinical perspective, this suggests that training solely on T may be sufficient for generalization, as the motion patterns in the sick dataset appear to be subsets of those in the healthy ones. To handle US videos with varying frame rates and lengths — originating from different probes and vendors — we apply frame skipping (0 to 5 frames) and extract video segments of varying lengths with approximately 50\% temporal overlap. To regularize the models for cyclic cardiac motion, we augment the reference motion by reversing entire sequences with a probability of $p = 0.2$ during training, effectively simulating backward tracking. All models are trained end-to-end using a weighted $L_1$ loss between the estimated and reference trajectories, following the approach of Zheng et al.~\cite{zheng2023pointodyssey}.

\section{Experiments and Results}
\subsection{Implementation details}
We implement all point tracking models in PyTorch based on their open-source repositories, unifying them for myocardium tissue tracking. A single training script with a learning rate of \(5 \times 10^{-4}\), a one-cycle scheduler~\cite{smith2019super}, and the AdamW optimizer ensures consistency. Due to limited resources, high fine-tuning cost (each over a week), and carbon footprint concerns, we decided to fully finetune only three SOTA point tracking models: PIPs++ (fast), EchoTracker (ultrasound-specific), and CoTracker3 (top performing) along with SpeckNet. EchoTracker was fine-tuned on dataset T for 35 epochs (8,620 clips, length 36, batch size 6), followed by one epoch on 1,922 full-length videos (batch size 1) with 45 points per sample. PIPs++ was fine-tuned on 8,618 clips (batch size 8) for 50 epochs, with the best weights at 12 epochs. CoTracker3 was fine-tuned on 8,620 clips (length 36, batch size 6) for 88 epochs, selecting 30 points per sample. Baselines for PIPs++ and EchoTracker were sourced from Azad et al.~\cite{azad2024echotracker}, and CoTracker3 was fine-tuned following the same configuration for 32 epochs, while the optimized weights emerged after 18 epochs. Finally, SpeckNet was trained on the same T dataset for 50 epochs (8,620 clips, length 36, batch size 4). Training and evaluation were performed on NVIDIA A6000 (48GB) and RTX 3090 (24GB) GPUs.

\subsection{Evaluation}
\subsubsection{Evaluation metrics.} Position accuracy (\(\delta^x\)) is the percentage of points within a pixel threshold from the ground truth, evaluated at \(x = 1, 2, 4, 8, 16\), and we report the average (\(\delta_{avg}^x\)) across these thresholds. Median trajectory error (MTE) measures the pixel distance between estimated and reference trajectories. We refer readers to the TAP-Vid paper~\cite{doersch2022tap} for details regarding these two matrices. Additionally, we report the average inference time (AIT) per video and peak GPU memory (PGM) usage on an RTX 3090 (24GB) GPU used for evaluation. For clinical relevance, we assess GLS — which quantifies the percentage change in the myocardial longitudinal length between ED and ES and compare it against expert reference measurements using the mean absolute deviation (MAD). The reference is obtained from a semi-automatic system provided by GE HealthCare.

\begin{figure}[h]
\centering
\includegraphics[width=\linewidth]{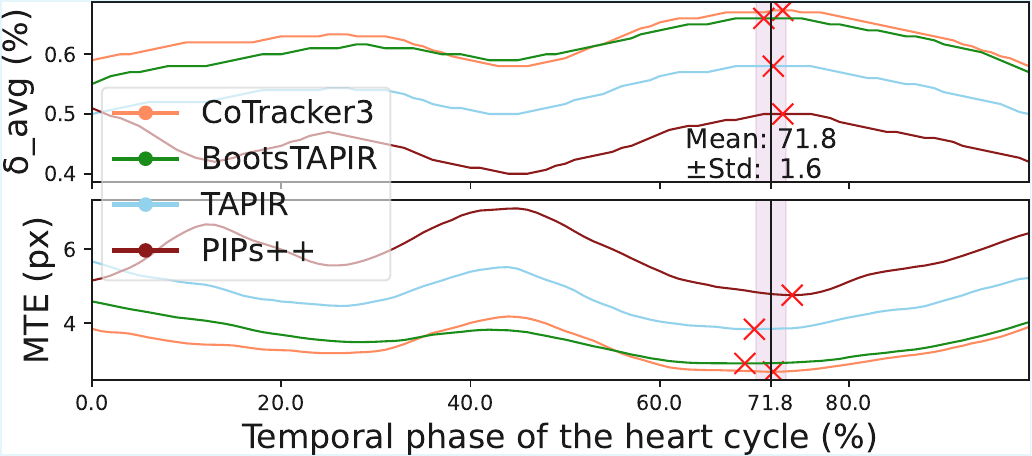}
\caption{Variation in tracking performance across temporal phases without fine-tuning. All models achieve the best position accuracy ($\delta^x$) and median trajectory error (MTE) at approximately 71.8\% into the cardiac cycle.} 
\label{fig:temp_perf_sota}
\end{figure}
\begin{table*}[ht]
\footnotesize
    \centering
    \caption{Comparison of tracking performance and efficiency between the re-fine-tuned models and their respective baselines on datasets D and E (OOD). Baseline values are shown in parentheses next to each result. Top-performing results are highlighted in bold, while the second-best are underlined.}
    \setlength{\tabcolsep}{1.5pt} 
    \begin{tabular}{lcccccccc|cccccccccc}
    \hline 
    \multirow{2}{*}{\textbf{Method}} & \multicolumn{8}{c}{\textbf{Dataset D (in-distribution)}} & \multicolumn{8}{c}{\textbf{Dataset E (OOD)}}\\
    \cline{2-17}
    & $<\delta^1\uparrow$ & $<\delta^2\uparrow$ & $<\delta^4\uparrow$ & $<\delta^8\uparrow$ & $<\delta^x_{avg}\uparrow$ & MTE$\downarrow$ & AIT$\downarrow$ & PGM$\downarrow$ & $<\delta^1\uparrow$ & $<\delta^2\uparrow$ & $<\delta^4\uparrow$ & $<\delta^8\uparrow$ & $<\delta^x_{avg}\uparrow$ & MTE$\downarrow$ & AIT$\downarrow$ & PGM$\downarrow$\\
    \hline
    PIPs++ & 26(15) & 52(39) & 83(73) & \underline{97}(95) & 72(65) & 2.23(3.0) & \textbf{0.16} & \textbf{1.77} & 21(9) & 40(23) & \underline{67}(49) & \underline{89}(76) & \underline{63}(50) & \underline{3.36}(5.74) & \textbf{0.16} & \textbf{1.80} &   \\
    EchoTracker & 28(22) & 54(48) & \underline{84}(80) & \underline{97}(96) & \underline{73}(69) & 2.22(2.61) & \underline{0.19} & 3.94 & 19(10) & 39(25) & 66(48) & 87(72) & 62(49) & 3.57(6.54) & \underline{0.19} & 4.02 & \\
    CoTracker3 & \textbf{33}(20) & \textbf{60}(45) & \textbf{87}(77) & \textbf{98}(95) & \textbf{76}(68) & \textbf{1.95}(2.8) & 0.63 & 2.45 & \textbf{26}(12) & \textbf{47}(30) & \textbf{72}(57) & \textbf{91}(81) & \textbf{67}(55) & \textbf{3.0}(4.87) & 0.63 & 2.50 & \\
    \hline
    SpeckNet (ours) & \underline{29} & \underline{55} & 83 & \underline{97} & \underline{73} & \underline{2.17} & 0.26 & \underline{1.89} & \underline{23} & \underline{43} & \underline{67} & 86 & \underline{63} & 3.48 & 0.25 & \underline{1.85} & \\
    \hline
    \end{tabular}
    \label{tab:track_perf}
    \\[-0.1cm]
    \raggedright
    {\scriptsize $<\delta^x_{avg}$: average position accuracy (\%), MTE: Median Trajectory Error (px), AIT: Average Inference Time (s), PGM: Peak GPU Memory (GB)}
\end{table*}
\subsubsection{Technical Results.}
We validate the choice of optimal temporal initialization by evaluating SOTA methods (without fine-tuning) across all temporal phases of the heart cycle in dataset D. As shown in Fig.~\ref{fig:temp_perf_sota}, we observe variations in both $\delta_{avg}^x$ and MTE across different phases. The optimal initialization point is identified at 71.9\% of the cardiac cycle, determined by averaging the phases that yield the best $\delta_{avg}^x$ and MTE across all models. This aligns with our hypothesis that diastasis offers the most stable period for initialization. We therefore use this optimal phase during inference to ensure maximal tracking performance. Furthermore, Fig.~\ref{fig:motion_dist} (bottom row) demonstrates that, applying our adopted affine transformations, the reference motions in dataset T become uniformly distributed in all directions across temporal phases — unlike in the original motion distribution shown in Fig.~\ref{fig:motion_dist} (top row). This augmentation preserves the underlying tissue structure and dynamics while eliminating directional bias. As a result, training on such impartial motion has significantly improved tracking performance across all evaluation metrics for the re-fine-tuned models compared to their baselines, on both in-distribution (D) and OOD (E) datasets, as reported in Table~\ref{tab:track_perf}. The results also reveal that CoTracker3 delivers the best tracking performance in echocardiography, while PIPs++ is the most efficient in terms of memory and computational cost. Notably, SpeckNet, despite being a purely spatial model, achieved the second-best performance, surpassing both PIPs++ and EchoTracker, highlighting its effectiveness as a domain-specific architecture. While all models outperformed their baselines, the improvement rate is higher on OOD data, indicating better generalization. 

\begin{figure}[h]
\centering
\includegraphics[width=\linewidth]{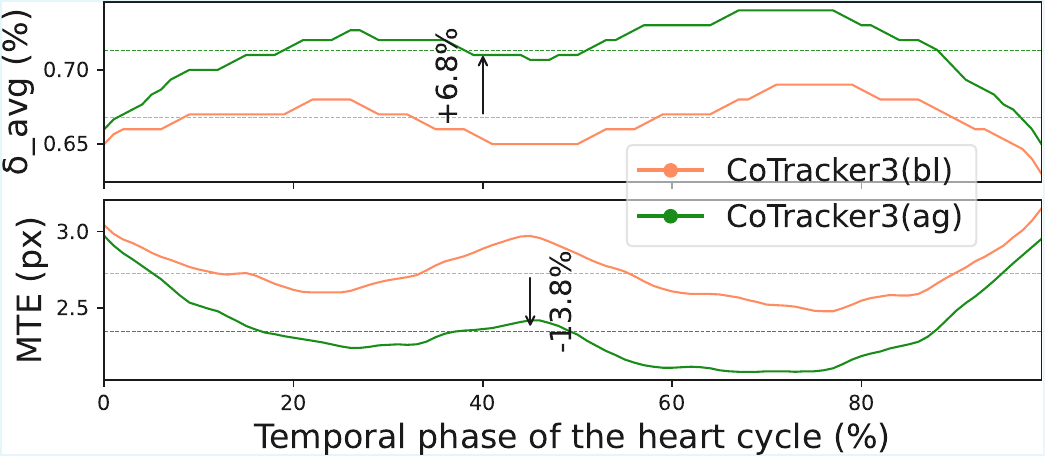}
\caption{The re-fine-tuned CoTracker3(ag) demonstrates improved positional accuracy (top) and reduced trajectory error (bottom) across all temporal phase initializations within the heart cycle compared to the baseline (bl).} 
\label{fig:temp_perf_cotracker3}
\end{figure}

To assess temporal invariance, we compare the tracking performance of the re-fine-tuned models, initialized from all temporal phases, against their respective baselines. Results for EchoTracker and CoTracker3 are shown in Fig.\ref{fig:paradigm} and \ref{fig:temp_perf_cotracker3} respectively. Although not perfectly invariant, the fine-tuned models show greater robustness and outperform baselines across most temporal positions. To highlight the improved tracking precision, we include $\delta^x$ at pixel thresholds of 1, 2, 4, and 8 in Table~\ref{tab:track_perf}. We also compute the average improvement across all models (CoTracker3, EchoTracker, and PIPs++) for each threshold. The gains are substantially higher at smaller thresholds compared to larger ones, with 55\% at 1px, 27\% at 2px, 10\% at 4px, and 2\% at 8px, on dataset D, respectively. Even greater improvements are observed on OOD data: 113\% at 1px, 62\% at 2px, 33\% at 4px, and 16\% at 8px. These results demonstrate enhanced tracking precision and generalization, both crucial for clinical application.

\subsubsection{Clinical Results.} 
The average GLS of the expert reference on the in-distribution (D) dataset was $(-13.5 \pm 5.2)\%$. The re-fine-tuned EchoTracker achieved a mean absolute deviation (MAD) of 1.3\% from the expert reference, which further improved to 1.2\% when point initialization was set at 72\% of the cardiac cycle. Notably, the best-performing model, CoTracker3, achieved a MAD of 1.1\% when initialized at cycle start and improved to 1.0\% at the 72\% temporal phase. These results align with the GLS agreement reported in the original EchoTracker study~\cite{azad2024echotracker}, while CoTracker3, to our knowledge, establishes a new SOTA in GLS estimation based on our technical analysis. 

\section{Conclusion \& Future work}
We have demonstrated that training advanced point tracking methods with impartial cardiac motion significantly enhances their robustness and generalization in echocardiography. Additionally, our carefully designed yet simple spatial model, \emph{SpeckNet}, outperformed several sophisticated models in US tracking, highlighting the benefit of domain-specific architectural choices. While we evaluated tracking generalizability on RV US data, we recommend against applying these models to such OOD scenarios in clinical settings without proper fine-tuning.

A notable limitation of our study lies in the reliance on human-annotated reference trajectories, which may not fully reflect the true ground truth. Future work will explore self- or weakly-supervised strategies to leverage unannotated data and incorporate temporal priors into SpeckNet to further improve its performance as a US-specific tracking model for echocardiography. While our overarching goal is to develop temporally invariant models, we find that optimal tracking still benefits from initialization at specific cardiac phases. Nonetheless, the re-fine-tuned models exhibit greater robustness and accuracy across all temporal phases compared to their respective baselines. Among all evaluated methods, CoTracker3 emerged as the top performer on both in-distribution and OOD datasets, while other models also demonstrated noticeable improvements. We believe that these robust, re-fine-tuned models will support the community in advancing various clinical analyses in echocardiography.

{
    \small
    \bibliographystyle{ieeenat_fullname}
    \bibliography{main}

\begin{thebibliography}{38}
\providecommand{\natexlab}[1]{#1}
\providecommand{\url}[1]{\texttt{#1}}
\expandafter\ifx\csname urlstyle\endcsname\relax
  \providecommand{\doi}[1]{doi: #1}\else
  \providecommand{\doi}{doi: \begingroup \urlstyle{rm}\Url}\fi

\bibitem[Angelini and Gerard(2006)]{angelini2006review}
Elsa~D Angelini and Olivier Gerard.
\newblock Review of myocardial motion estimation methods from optical flow tracking on ultrasound data.
\newblock In \emph{2006 International Conference of the IEEE Engineering in Medicine and Biology Society}, pages 1537--1540. IEEE, 2006.

\bibitem[Aydemir et~al.(2025)Aydemir, Cai, Xie, and G\"uney]{Aydemir2025ICLR}
G\"orkay Aydemir, Xiongyi Cai, Weidi Xie, and Fatma G\"uney.
\newblock {Track-On}: Transformer-based online point tracking with memory.
\newblock In \emph{The Thirteenth International Conference on Learning Representations}, 2025.

\bibitem[Azad et~al.(2024)Azad, Chernyshov, Nyberg, Tveten, Lovstakken, Dalen, Grenne, and {\O}stvik]{azad2024echotracker}
Md~Abulkalam Azad, Artem Chernyshov, John Nyberg, Ingrid Tveten, Lasse Lovstakken, H{\aa}vard Dalen, Bj{\o}rnar Grenne, and Andreas {\O}stvik.
\newblock Echotracker: Advancing myocardial point tracking in echocardiography.
\newblock \emph{arXiv preprint arXiv:2405.08587}, 2024.

\bibitem[Chen et~al.(2024)Chen, Schmidt, Prisman, and Salcudean]{chen2024pipsus}
Wanwen Chen, Adam Schmidt, Eitan Prisman, and Septimiu~E Salcudean.
\newblock Pipsus: Self-supervised point tracking in ultrasound.
\newblock In \emph{International Workshop on Advances in Simplifying Medical Ultrasound}, pages 47--57. Springer, 2024.

\bibitem[Chernyshov et~al.(2025)Chernyshov, Nyberg, Holmstr{\o}m, Azad, Grenne, Dalen, Aase, Lovstakken, and {\O}stvik]{chernyshov2025low}
Artem Chernyshov, John Nyberg, Vegard Holmstr{\o}m, Md~Abulkalam Azad, Bj{\o}rnar Grenne, H{\aa}vard Dalen, Svein~Arne Aase, Lasse Lovstakken, and Andreas {\O}stvik.
\newblock Low complexity point tracking of the myocardium in 2d echocardiography.
\newblock \emph{arXiv preprint arXiv:2503.10431}, 2025.

\bibitem[Cho et~al.(2024)Cho, Huang, Nam, An, Kim, and Lee]{cho2024local}
Seokju Cho, Jiahui Huang, Jisu Nam, Honggyu An, Seungryong Kim, and Joon-Young Lee.
\newblock Local all-pair correspondence for point tracking.
\newblock In \emph{European Conference on Computer Vision}, pages 306--325. Springer, 2024.

\bibitem[Deng et~al.(2022)Deng, Cai, Zhang, Cao, Chen, Jiang, Zhuang, and Wang]{deng2022myocardial}
Yinlong Deng, Peiwei Cai, Li Zhang, Xiongcheng Cao, Yequn Chen, Shiyan Jiang, Zhemin Zhuang, and Bin Wang.
\newblock Myocardial strain analysis of echocardiography based on deep learning.
\newblock \emph{Frontiers in Cardiovascular Medicine}, 9:\penalty0 1067760, 2022.

\bibitem[D'hooge(2007)]{d2007principles}
Jan D'hooge.
\newblock Principles and different techniques for speckle tracking.
\newblock \emph{Myocardial imaging: tissue Doppler and speckle tracking}, pages 17--25, 2007.

\bibitem[Doersch et~al.(2022)Doersch, Gupta, Markeeva, Recasens, Smaira, Aytar, Carreira, Zisserman, and Yang]{doersch2022tap}
Carl Doersch, Ankush Gupta, Larisa Markeeva, Adri{\`a} Recasens, Lucas Smaira, Yusuf Aytar, Jo{\~a}o Carreira, Andrew Zisserman, and Yi Yang.
\newblock Tap-vid: A benchmark for tracking any point in a video.
\newblock \emph{Advances in Neural Information Processing Systems (NeurIPS)}, 35:\penalty0 13610--13626, 2022.

\bibitem[Doersch et~al.(2023)Doersch, Yang, Vecerik, Gokay, Gupta, Aytar, Carreira, and Zisserman]{doersch2023tapir}
Carl Doersch, Yi Yang, Mel Vecerik, Dilara Gokay, Ankush Gupta, Yusuf Aytar, Joao Carreira, and Andrew Zisserman.
\newblock Tapir: Tracking any point with per-frame initialization and temporal refinement.
\newblock \emph{ICCV}, 2023.

\bibitem[Doersch et~al.(2024)Doersch, Luc, Yang, Gokay, Koppula, Gupta, Heyward, Rocco, Goroshin, Carreira, et~al.]{doersch2024bootstap}
Carl Doersch, Pauline Luc, Yi Yang, Dilara Gokay, Skanda Koppula, Ankush Gupta, Joseph Heyward, Ignacio Rocco, Ross Goroshin, Jo{\~a}o Carreira, et~al.
\newblock Bootstap: Bootstrapped training for tracking-any-point.
\newblock In \emph{Proceedings of the Asian Conference on Computer Vision}, pages 3257--3274, 2024.

\bibitem[Dosovitskiy et~al.(2015)Dosovitskiy, Fischer, Ilg, Hausser, Hazirbas, Golkov, Van Der~Smagt, Cremers, and Brox]{dosovitskiy2015flownet}
Alexey Dosovitskiy, Philipp Fischer, Eddy Ilg, Philip Hausser, Caner Hazirbas, Vladimir Golkov, Patrick Van Der~Smagt, Daniel Cremers, and Thomas Brox.
\newblock Flownet: Learning optical flow with convolutional networks.
\newblock In \emph{Proceedings of the IEEE international conference on computer vision}, pages 2758--2766, 2015.

\bibitem[Evain et~al.(2022)Evain, Sun, Faraz, Garcia, Saloux, Gerber, De~Craene, and Bernard]{evain2022motion}
Ewan Evain, Yunyun Sun, Khuram Faraz, Damien Garcia, Eric Saloux, Bernhard~L Gerber, Mathieu De~Craene, and Olivier Bernard.
\newblock Motion estimation by deep learning in 2d echocardiography: synthetic dataset and validation.
\newblock \emph{IEEE transactions on medical imaging}, 41\penalty0 (8):\penalty0 1911--1924, 2022.

\bibitem[Harley et~al.(2022)Harley, Fang, and Fragkiadaki]{harley2022particle}
Adam~W Harley, Zhaoyuan Fang, and Katerina Fragkiadaki.
\newblock Particle video revisited: Tracking through occlusions using point trajectories.
\newblock In \emph{European Conference on Computer Vision (ECCV)}, pages 59--75. Springer, 2022.

\bibitem[Horn and Schunck(1981)]{horn1981determining}
Berthold~KP Horn and Brian~G Schunck.
\newblock Determining optical flow.
\newblock \emph{Artificial intelligence}, 17\penalty0 (1-3):\penalty0 185--203, 1981.

\bibitem[Ilg et~al.(2017)Ilg, Mayer, Saikia, Keuper, Dosovitskiy, and Brox]{ilg2017flownet}
Eddy Ilg, Nikolaus Mayer, Tonmoy Saikia, Margret Keuper, Alexey Dosovitskiy, and Thomas Brox.
\newblock Flownet 2.0: Evolution of optical flow estimation with deep networks.
\newblock In \emph{Proceedings of the IEEE conference on computer vision and pattern recognition (CVPR)}, pages 2462--2470, 2017.

\bibitem[Joos et~al.(2018)Joos, Por{\'e}e, Liebgott, Vray, Baudet, Faurie, Tournoux, Cloutier, Nicolas, and Garcia]{joos2018high}
Philippe Joos, Jonathan Por{\'e}e, Herv{\'e} Liebgott, Didier Vray, Mathilde Baudet, Julia Faurie, Fran{\c{c}}ois Tournoux, Guy Cloutier, Barbara Nicolas, and Damien Garcia.
\newblock High-frame-rate speckle-tracking echocardiography.
\newblock \emph{IEEE transactions on ultrasonics, ferroelectrics, and frequency control}, 65\penalty0 (5):\penalty0 720--728, 2018.

\bibitem[Karaev et~al.(2024{\natexlab{a}})Karaev, Makarov, Wang, Neverova, Vedaldi, and Rupprecht]{karaev2024cotracker3}
Nikita Karaev, Iurii Makarov, Jianyuan Wang, Natalia Neverova, Andrea Vedaldi, and Christian Rupprecht.
\newblock {CoTracker3}: Simpler and better point tracking by pseudo-labelling real videos.
\newblock 2024{\natexlab{a}}.

\bibitem[Karaev et~al.(2024{\natexlab{b}})Karaev, Rocco, Graham, Neverova, Vedaldi, and Rupprecht]{karaev2023cotracker}
Nikita Karaev, Ignacio Rocco, Benjamin Graham, Natalia Neverova, Andrea Vedaldi, and Christian Rupprecht.
\newblock Cotracker: It is better to track together.
\newblock In \emph{European Conference on Computer Vision (ECCV)}, 2024{\natexlab{b}}.

\bibitem[Li et~al.(2024)Li, Zhang, Liu, Zeng, Ren, Li, and Zhang]{li2024taptr}
Hongyang Li, Hao Zhang, Shilong Liu, Zhaoyang Zeng, Tianhe Ren, Feng Li, and Lei Zhang.
\newblock Taptr: Tracking any point with transformers as detection.
\newblock In \emph{European Conference on Computer Vision}, pages 57--75. Springer, 2024.

\bibitem[Mailloux et~al.(1987)Mailloux, Bleau, Bertrand, and Petitclerc]{mailloux1987computer}
Guy~E Mailloux, Andre Bleau, Michel Bertrand, and Robert Petitclerc.
\newblock Computer analysis of heart motion from two-dimensional echocardiograms.
\newblock \emph{IEEE Transactions on Biomedical Engineering}, \penalty0 (5):\penalty0 356--364, 1987.

\bibitem[Meunier et~al.(1988)Meunier, Bertrand, Mailloux, and Petitclerc]{meunier1988local}
J Meunier, M Bertrand, GE Mailloux, and R Petitclerc.
\newblock Local myocardial deformation computed from speckle motion.
\newblock In \emph{Proceedings. Computers in Cardiology 1988}, pages 133--136. IEEE, 1988.

\bibitem[{\O}stvik et~al.(2018){\O}stvik, Smistad, Espeland, Berg, and Lovstakken]{ostvik2018automatic}
Andreas {\O}stvik, Erik Smistad, Torvald Espeland, Erik Andreas~Rye Berg, and Lasse Lovstakken.
\newblock Automatic myocardial strain imaging in echocardiography using deep learning.
\newblock In \emph{International Workshop on Deep Learning in Medical Image Analysis}, pages 309--316. Springer, 2018.

\bibitem[{\O}stvik et~al.(2021){\O}stvik, Salte, Smistad, Nguyen, Melichova, Brunvand, Haugaa, Edvardsen, Grenne, and Lovstakken]{ostvik2021myocardial}
Andreas {\O}stvik, Ivar~Mj{\aa}land Salte, Erik Smistad, Thuy~Mi Nguyen, Daniela Melichova, Harald Brunvand, Kristina Haugaa, Thor Edvardsen, Bj{\o}rnar Grenne, and Lasse Lovstakken.
\newblock Myocardial function imaging in echocardiography using deep learning.
\newblock \emph{ieee transactions on medical imaging}, 40\penalty0 (5):\penalty0 1340--1351, 2021.

\bibitem[Pollock and Makaryus(2023)]{pollok2023physiology}
Joshua~D. Pollock and Amgad~N. Makaryus.
\newblock \emph{Physiology, Cardiac Cycle}.
\newblock StatPearls Publishing, Treasure Island (FL), 2023.

\bibitem[Por{\'e}e et~al.(2018)Por{\'e}e, Baudet, Tournoux, Cloutier, and Garcia]{poree2018dual}
Jonathan Por{\'e}e, Mathilde Baudet, Fran{\c{c}}ois Tournoux, Guy Cloutier, and Damien Garcia.
\newblock A dual tissue-doppler optical-flow method for speckle tracking echocardiography at high frame rate.
\newblock \emph{IEEE transactions on medical imaging}, 37\penalty0 (9):\penalty0 2022--2032, 2018.

\bibitem[Salte et~al.(2021)Salte, Østvik, Smistad, Melichova, Nguyen, Karlsen, Brunvand, Haugaa, Edvardsen, Lovstakken, and Grenne]{salte2021aistrain}
Ivar~M. Salte, Andreas Østvik, Erik Smistad, Daniela Melichova, Thuy~Mi Nguyen, Sigve Karlsen, Harald Brunvand, Kristina~H. Haugaa, Thor Edvardsen, Lasse Lovstakken, and Bjørnar Grenne.
\newblock Artificial intelligence for automatic measurement of left ventricular strain in echocardiography.
\newblock \emph{JACC: Cardiovascular Imaging}, 14\penalty0 (10):\penalty0 1918--1928, 2021.

\bibitem[Sand and Teller(2008)]{sand2008particle}
Peter Sand and Seth Teller.
\newblock Particle video: Long-range motion estimation using point trajectories.
\newblock \emph{International journal of computer vision}, 80:\penalty0 72--91, 2008.

\bibitem[Shen et~al.(2024)Shen, Zhu, Zhou, Liu, Yi, Dong, Zhao, Brady, Cao, Ma, et~al.]{shen2024continuous}
Chengkang Shen, Hao Zhu, You Zhou, Yu Liu, Si Yi, Lili Dong, Weipeng Zhao, David~J Brady, Xun Cao, Zhan Ma, et~al.
\newblock Continuous 3d myocardial motion tracking via echocardiography.
\newblock \emph{IEEE Transactions on Medical Imaging}, 2024.

\bibitem[Shi et~al.(2016)Shi, Caballero, Husz{\'a}r, Totz, Aitken, Bishop, Rueckert, and Wang]{shi2016real}
Wenzhe Shi, Jose Caballero, Ferenc Husz{\'a}r, Johannes Totz, Andrew~P Aitken, Rob Bishop, Daniel Rueckert, and Zehan Wang.
\newblock Real-time single image and video super-resolution using an efficient sub-pixel convolutional neural network.
\newblock In \emph{Proceedings of the IEEE conference on computer vision and pattern recognition}, pages 1874--1883, 2016.

\bibitem[Smith and Topin(2019)]{smith2019super}
Leslie~N Smith and Nicholay Topin.
\newblock Super-convergence: Very fast training of neural networks using large learning rates.
\newblock In \emph{Artificial intelligence and machine learning for multi-domain operations applications}, pages 369--386. SPIE, 2019.

\bibitem[Suhling et~al.(2005)Suhling, Arigovindan, Jansen, Hunziker, and Unser]{suhling2005myocardial}
Michael Suhling, Muthuvel Arigovindan, Christian Jansen, Patrick Hunziker, and Michael Unser.
\newblock Myocardial motion analysis from b-mode echocardiograms.
\newblock \emph{IEEE Transactions on image processing}, 14\penalty0 (4):\penalty0 525--536, 2005.

\bibitem[Sun et~al.(2018)Sun, Yang, Liu, and Kautz]{sun2018pwc}
Deqing Sun, Xiaodong Yang, Ming-Yu Liu, and Jan Kautz.
\newblock Pwc-net: Cnns for optical flow using pyramid, warping, and cost volume.
\newblock In \emph{Proceedings of the IEEE conference on computer vision and pattern recognition (CVPR)}, pages 8934--8943, 2018.

\bibitem[Trahey et~al.(1988)Trahey, Hubbard, and von Ramm]{trahey1988angle}
Gregg~E Trahey, SM Hubbard, and Olaf~T von Ramm.
\newblock Angle independent ultrasonic blood flow detection by frame-to-frame correlation of b-mode images.
\newblock \emph{Ultrasonics}, 26\penalty0 (5):\penalty0 271--276, 1988.

\bibitem[Tumanyan et~al.(2024)Tumanyan, Singer, Bagon, and Dekel]{tumanyan2024dino}
Narek Tumanyan, Assaf Singer, Shai Bagon, and Tali Dekel.
\newblock Dino-tracker: Taming dino for self-supervised point tracking in a single video.
\newblock In \emph{European Conference on Computer Vision}, pages 367--385. Springer, 2024.

\bibitem[Vaswani et~al.(2017)Vaswani, Shazeer, Parmar, Uszkoreit, Jones, Gomez, Kaiser, and Polosukhin]{vaswani2017attention}
Ashish Vaswani, Noam Shazeer, Niki Parmar, Jakob Uszkoreit, Llion Jones, Aidan~N Gomez, {\L}ukasz Kaiser, and Illia Polosukhin.
\newblock Attention is all you need.
\newblock \emph{Advances in neural information processing systems}, 30, 2017.

\bibitem[Voigt et~al.(2015)Voigt, Pedrizzetti, Lysyansky, Marwick, Houle, Baumann, Pedri, Ito, Abe, Metz, et~al.]{voigt2015definitions}
Jens-Uwe Voigt, Gianni Pedrizzetti, Peter Lysyansky, Tom~H Marwick, Helen Houle, Rolf Baumann, Stefano Pedri, Yasuhiro Ito, Yasuhiko Abe, Stephen Metz, et~al.
\newblock Definitions for a common standard for 2d speckle tracking echocardiography: consensus document of the eacvi/ase/industry task force to standardize deformation imaging.
\newblock \emph{European Heart Journal-Cardiovascular Imaging}, 16\penalty0 (1):\penalty0 1--11, 2015.

\bibitem[Zheng et~al.(2023)Zheng, Harley, Shen, Wetzstein, and Guibas]{zheng2023pointodyssey}
Yang Zheng, Adam~W Harley, Bokui Shen, Gordon Wetzstein, and Leonidas~J Guibas.
\newblock Pointodyssey: A large-scale synthetic dataset for long-term point tracking.
\newblock In \emph{Proceedings of the IEEE/CVF International Conference on Computer Vision (ICCV)}, pages 19855--19865, 2023.

\end{thebibliography}
}

\end{document}